\documentclass[lettersize,journal]{IEEEtran}
\usepackage{graphics}
\usepackage{amsmath}
\usepackage{amsfonts}
\usepackage{colortbl}
\usepackage{amssymb} 
\usepackage{graphicx}
\usepackage{subfig}
\usepackage{hyperref}
\usepackage[sort,compress]{cite}

\usepackage{lipsum}
\usepackage{balance}
\usepackage{dblfloatfix}    
\usepackage{booktabs}
\usepackage{tikz}

\usepackage{arydshln} 

\usepackage{enumitem} 

\newcommand{\realfield}{\hbox{I \kern -.4em R}}
\newcommand {\mb}[1]{\mathbf{#1}} 
\newcommand {\bs}[1]{\boldsymbol{#1}}

\usepackage{soul} 

\usepackage{booktabs}
\usepackage{graphicx}

\graphicspath{figures/pdf/}

\begin{document}
%
\title{A High-Fidelity Simulation Framework for Grasping Stability Analysis in Human Casualty Manipulation}
%
%
%
%
%
%
\author{Qianwen~Zhao$^{1}$, Rajarshi~Roy$^{2}$, Chad~Spurlock$^{2}$, Kevin~Lister$^{2}$, and~Long~Wang$^{1}$
\thanks{
	$^{1}$Qianwen Zhao and Long Wang are with Charles V. Schaefer, Jr.School of Engineering and Science, Department of Mechanical Engineering, Stevens Institute of Technology, Hoboken, New Jersey, USA 07030
	{\tt\footnotesize qzhao10, lwang4@stevens.edu}
}%
\thanks{
	$^{2}$ Rajarshi~Roy, Chad~Spurlock, and Kevin~Lister are with Corvid Technologies, Mooresville, NC, USA 28117
	{\tt\footnotesize rajarshi.roy, chad.spurlock, kevin.lister@corvidtec.com}
}
}

\maketitle
\begin{abstract}
Recently, there has been a growing interest in rescue robots due to their vital role in addressing emergency scenarios and providing crucial assistance in challenging or hazardous situations where human intervention is problematic. 
However, very few of these robots are capable of actively engaging with humans and undertaking physical manipulation tasks. 
This limitation is largely attributed to the absence of tools that can realistically simulate physical interactions, especially the contact mechanisms between a robotic gripper and a human body. 
In this study, we aim to address key limitations in current developments towards robotic casualty manipulation. 
Firstly, we present an integrative simulation framework for casualty manipulation. We adapt a finite element method (FEM) tool into the grasping and manipulation scenario, and the developed framework can provide accurate biomechanical reactions resulting from manipulation. 
Secondly, we conduct a detailed assessment of grasping stability during casualty grasping and manipulation simulations. To validate the necessity and superior performance of the proposed high-fidelity simulation framework, we conducted a qualitative and quantitative comparison of grasping stability analyses between the proposed framework and the state-of-the-art multi-body physics simulations.
Through these efforts, we have taken the first step towards a feasible solution for robotic casualty manipulation.\par

\end{abstract}

\section{Introduction}

\begin{figure}[!t]
	\centering
	\includegraphics[width=0.99\columnwidth]{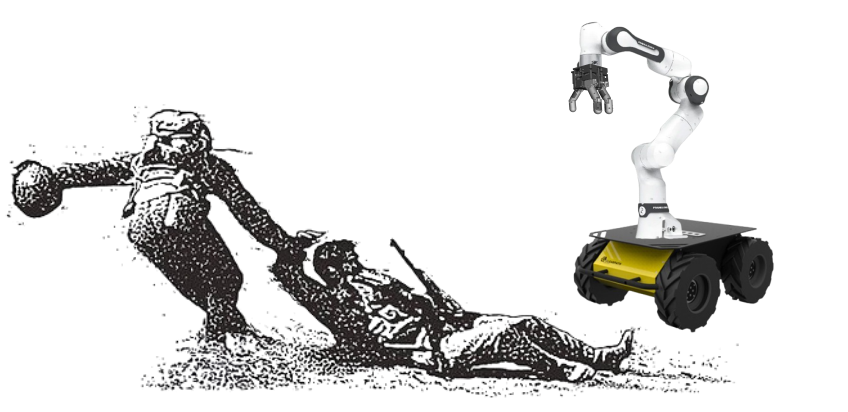}
	\caption{The prospect for using autonomous robot systems in casualty manipulation.}
	\label{fig:casualty-manipulation}
\end{figure}

The use of robotics and autonomous systems (RAS) in search-and-rescue missions has gained increasingly attention recently \cite{adams2013robotic,adams2024human,murphy2007search}. These systems have the potential to significantly enhance the capabilities of medical teams. A conceptual drawing of the use case of such RAS is shown in Fig.~\ref{fig:casualty-manipulation}, where the RAS extracts a wounded soldier from a contested area. Effective human-robot interaction (HRI) systems are crucial in these applications, as RAS must cooperate seamlessly with human rescue teams. These systems must also handle environmental challenges while ensuring the safe and efficient grasping and manipulation of incapacitated soldiers. Achieving this requires a thorough understanding of potential injury mechanisms in HRI. Therefore, planning and control software should incorporate a biomechanically accurate human model to inform real-time autonomous planning decisions. \par

\par

However, currently available technology is far from sufficient in providing the aforementioned capabilities. Most robot manipulation systems involve three primary steps to achieve a task: (i) perception for understanding the environment; (ii) planning safe trajectories; and (iii) execution of the planned trajectories. Even if we set aside (i) and (iii), assuming the RAS has complete knowledge of its surroundings and the full ability to execute the planned motion, it is still currently incapable of (ii) planning safe and complex casualty manipulations. This limitation is due to the lack of high-fidelity human models and simulation environments.\par

\begin{figure*}[!t]
	\centering
	\includegraphics[width=1.99\columnwidth]{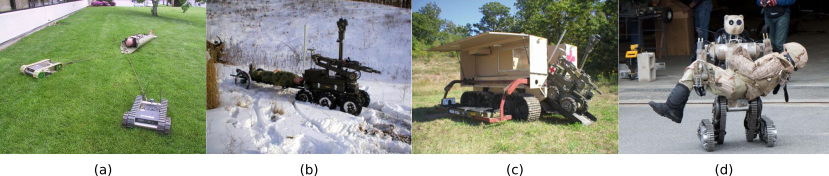}
	\caption{A summary of current casualty extraction robots. (a) Valkyrie \cite{ValkyrieSBIR} is one of the earliest investigations on using mobile robot for the recovery of battlefield casualties. (b) REX \cite{REXREVsbir} approaches the victim with wheeled stretcher to load and drag the victim back to the larger combat medical vehicle (REV \cite{REXREVsbir}) in (c). (d) The Battlefield Extraction Assist Robot (BEAR \cite{BEAR}) is the first humanoid robot in robot assisted extraction and evacuation. It consists of a humanoid hydraulic torso and a mobile robot base, and it can carry the victim up to 500 pounds. }
    \label{fig:casualty-extraction-robots}
\end{figure*}

Most of the literature in the field of search-and-rescue focuses on the discovery and localization of victims, visual assessment of victims, and environmental information gathering \cite{Bhatia2011,Niroui2019,Lygouras2019,Cardona2019,Deng2020, Queralta2020,Cruz2021,micheal_yip_contactless_weight}.
These robots do not have physical interactions with victims.  
Some rescue robots focus on transporting victims through unstructured terrains \cite{ValkyrieSBIR,REXREVsbir,BookSurgicalRobotics}.  
Few of them consider active physical contact with humans. For instance, \cite{BEAR,Sun2022,Yim2009,yim2011towards,williams2017analysis,williams2019robotic} demonstrate the ability to manipulate human bodies without considering the biomechanical reactions of the human body. 
These types of robots require much more careful motion planning (step (ii)) since they involve physical interactions with the victim. A summary of current casualty extraction robots is presented in Fig.~\ref{fig:casualty-extraction-robots}. However, current models treat humans merely as objects, indistinguishable from sandbags.
Most recently, research has advanced in generating biomechanically safe trajectories for search-and-rescue missions, as detailed in \cite{micheal_yip}. This research takes into account human modeling as a series of rigid links interconnected by joints. \par

\par

However, the modeling of humans in the above examples is very preliminary. These models do not differentiate between muscles, bones, organs, and soft tissues, and biomechanical reactions are not considered. Research focused on simulating human soft tissues for surgical applications does exist. For example, \cite{FEM_soft_tissue_liver} developed a soft liver model using finite element methods (FEM) to support surgery planning and training, while \cite{Palmeri2010} presented soft tissue models using FEM and studied their response to impulsive acoustic radiation force. More comprehensive models of human soft tissues for surgical applications can be found in survey papers \cite{FEM_soft_tissue_survey_2005, FEM_soft_tissue_survey_2019}. Additionally, research on modeling soft contacts \cite{MatieGraspWithSoftFingerDeformation, IPC_GraspSim, Xu2021TRO} has been undertaken. However, none of the above presents a comprehensive high-fidelity model of the entire human body capable of assisting in casualty manipulation. The challenges of casualty manipulation remain.\par

Therefore, in addressing critical limitations in human modeling and the grasping problem for robot-assisted casualty manipulation, the contributions of this work are summarized below.
\begin{itemize}
\item[1.] We developed a novel casualty manipulation simulation framework by utilizing a high-fidelity digital human model within an explicit dynamic FEM framework. This framework carefully models the human musculoskeletal system, including skin, fascia, fat, and ligaments. It achieves accurate simulation of human biomechanical reactions resulting from robot manipulation actions, filling the absence of such a high-fidelity tool in the field.
\item[2.] Enabled by the developed framework, we investigated the grasping stability problem on casualty limbs. We provided qualitative and quantitative results using the proposed framework and compared them with state-of-the-art (SotA) casualty manipulation simulation methods. The findings revealed limitations and unrealistic aspects of the SotA casualty manipulation simulation methods.
\end{itemize}

To the best of our knowledge, we are the first to introduce FEM into the simulation pipeline for robotic manipulation of high-fidelity digital human models. The human models in this work consider muscles as passive but have the potential to be extended to include active muscle activation functions in the future. Furthermore, the analysis of grasping stability lays the foundation for the successful execution of casualty manipulation tasks.\par

\section{High-fidelity Simulation Framework for Casualty Manipulation}

The proposed framework enables high-fidelity simulations of casualty manipulation by (i) introducing FEM and a digital human model to the pipeline, and (ii) bridging the gap between physics-based multi-body dynamics simulators and finite-element (FE) solvers.
The integration of these two types of simulators provides a comprehensive platform that combines the efficiency of the former in robot system kinematics computation with the high-fidelity soft contact mechanism of the latter. 
This synthesis opens the door to a wide range of applications, with a particular emphasis on scenarios involving robotic manipulation of human-like entities.\par

Conventional robotics simulations, such as those using tools like Gazebo and MuJoCo, excel in representing rigid body interactions and are widely used for robot motion planning. However, robustness in a simulation does not necessarily equate to realism. These tools fall short when handling soft contacts and deformations, particularly in scenarios involving human beings or soft objects. When grasping an object, the contacts between the object and a human hand are never just point contacts. In real-life scenarios, it is impossible for only one point of each finger to be in contact with the object. Instead, contact will always result in the deformation of the human hand muscles, the object, or both. Contacts in daily life are always surface contacts, involving a large area of each finger coming into contact with the object. Conventional robotics simulators often fail to accurately simulate this type of contact.

On the other side of the simulators' spectrum, FE solvers, designed to handle continuous media, offer a more detailed representation of intra-system behaviors. However, they struggle with the complexities of robot system kinematics due to breaking each robot component into individual nodes. Therefore, a complementary integration of the two fills each other's gaps. It enables high-fidelity simulations that require both precise robot grasping control and accurate modeling of contacts with deformable human bodies.\par

Below, we discuss grasps in both multi-body dynamics and FEM environments, respectively.\par

\subsection{Equilibrium grasps in multi-body dynamics simulators} \label{subsec:grasp-in-mj}

Simulations of grasping in multi-body dynamics simulators usually depend on rigid body mechanics, where both the robotic hand and the grasped object are modeled as rigid bodies.
When two objects come into contact, the physics engine calculates the contact forces $\boldsymbol{\lambda}_c$ that act at the contact points. These forces include normal forces, $\mathbf{f}_n$, and frictional forces, $\mathbf{f}_t$. For a given normal force $\mathbf{f}_n$, the frictional force $\mathbf{f}_t$ is limited by Coulomb's model to prevent relative motion between the contacting points: 
\begin{equation}
\boldsymbol{\lambda}_c = \mathbf{f}_n + \mathbf{f}_t,
\quad
|\mathbf{f}_t| \leq \ \mu \ |\mathbf{f}_n|,
\quad
\alpha = \tan^{-1} \mu
\end{equation}
where $\mu$ is the friction coefficient at the contact location, and $\alpha$ is the friction angle.\par

\begin{figure}[!t]
	\centering
	\vspace{1mm}
	\includegraphics[width=0.99\columnwidth]{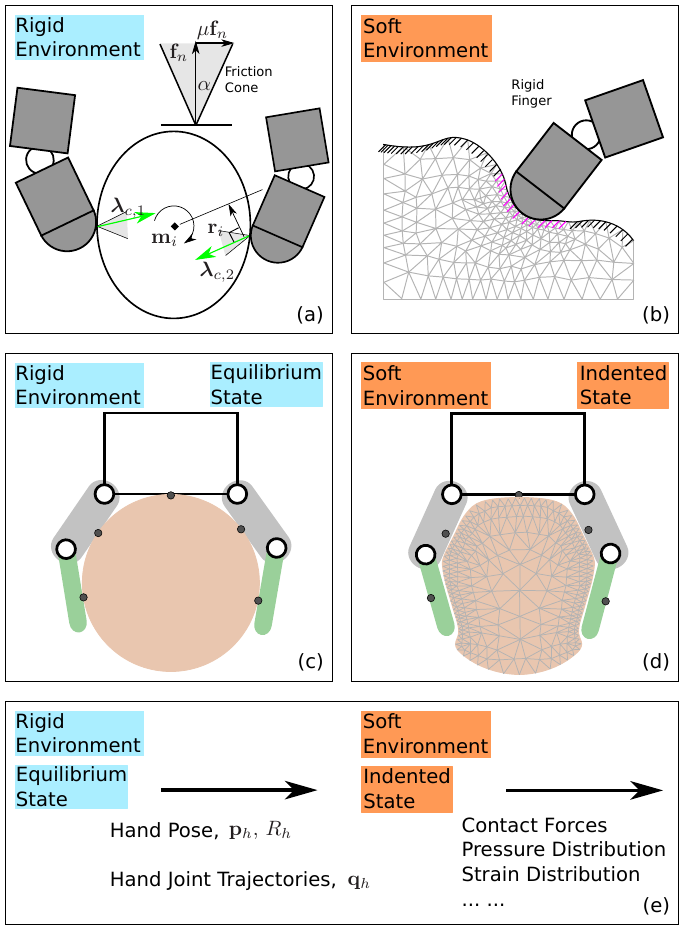}
	\caption{A comparison of grasping in a SotA multi-body dynamics simulator (the rigid environment) and in an FE solver (soft environment). (a) (c) illustrate rigid point contacts, and (b) (d) demonstrate their soft environment counterparts. (e) is a simplified data flow diagram of the proposed simulation framework. }
	\label{fig:rigid-vs-soft}
\end{figure} \par

Consider an object grasped with $N$ point contacts. 
Let $\boldsymbol{\lambda}_{c,i}$ denotes the contact force at the $i^\text{th}$ contact location. 
Shown in Fig.~\ref{fig:rigid-vs-soft}(a), the grasping static equilibrium is governed by:

\begin{subequations} \label{eq:force_moment_eq}
\begin{align}
    &\sum_{i=1}^{N} \boldsymbol{\lambda}_{c,i} + m\mathbf{g} = \mb{0} \\
    &\sum_{i=1}^{N} (\mathbf{m}_i +  \mathbf{r}_i \times \boldsymbol{\lambda}_{c,i}) = \mb{0} \\
    &\boldsymbol{\lambda}_{c,i}, \mb{m}_i \in \realfield^3
    \end{align}
\end{subequations}
where $m\mathbf{g}$ represents gravity, and 
$\mb{m}_i$, $\mb{r}_i$ denote frictional moment, and moment arm, at the $i^\text{th}$ contact, respectively.\par

When resolving contacts during grasping, the existence of external forces (such as gravity) causes the physics engine to decompose the external force into components parallel and perpendicular to the contact surface. It then checks if the parallel component is within the friction cone (as illustrated in Fig.~\ref{fig:rigid-vs-soft}~(a)). The object remains fixed if the frictional force is sufficient to sustain a fixed contact; otherwise, it transitions to motion. By continuously evaluating the forces and constraints at contact points and comparing them against the friction cone, the engines effectively simulate object interactions.\par

Figure~\ref{fig:rigid-vs-soft}~(c) illustrates the grasping \textit{Equilibrium State} in a rigid simulation environment where the object does not undergo translation or rotation. \par

\begin{figure}[!b]
	\centering
	\includegraphics[width=0.45\columnwidth]{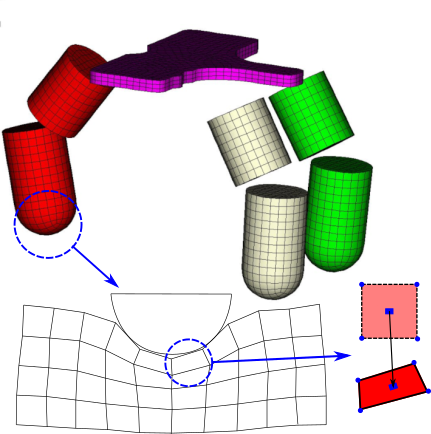}
	\caption{Illustration of soft contact interactions in an FE solver, including definitions of nodes and elements.}
	\label{fig:MFRH_mesh}
\end{figure}\par

\subsection{Soft contacts equilibrium grasps using FEM}\label{subsec:grasp-in-fem}


FEM is a numerical technique used to solve partial differential equations across discrete domains consisting of nodes and elements. It is applied to complex problems in areas such as deformable structual analysis, heat transfer, and fluid dynamics. Despite its potential, the applicaiton of FEM in robotics remains under-explored, possibly due to the computational challenges associated with discretizing large domains.
In robotics, FEM is particularly effective for accurately modeling soft contacts and human limb manipulation, which involve interconnected parts with nonlinear and heterogeneous properties.
It provides clear advantages over other methods, including (i) multi-body dynamics simulators (as discussed in section~\ref{subsec:grasp-in-mj}), (ii) simplified analytical modeling techniques such as pseudo-rigid-body models \cite{howell2013compliant}, and (iii) large deflection beam analysis using elliptical integrals \cite{faybook}.
Opposed to the rigid body environment shown in Fig.~\ref{fig:rigid-vs-soft}~(a,c), Fig.~\ref{fig:rigid-vs-soft}~(b,d) illustrates the interaction between a rigid robotic hand and a soft object.
In Fig.~\ref{fig:rigid-vs-soft}~(d), the soft object deforms under contact forces. This grasping condition is referred as \textit{Indented State} in this paper. 
We will briefly review the basic formulations of explicit dynamic nonlinear FEM, focusing on the \textit{Governing Equations}, \textit{Contact Modeling}, and \textit{Strain Calculations for Deformations}, which are crucial for understanding soft contacts and grasping stability.

\subsubsection{Governing Equations}
For a discretized domain comprising $m$ disjointed elements and $k$ nodal points defining each element, the discrete equations of motion for explicit dynamic nonlinear FEM can be derived from the momentum balance equation. For example, in a simplified planar illustration in Fig.~\ref{fig:MFRH_mesh}, the soft object has 40 elements, and each element is defined by 4 nodal points. The equations are summarized below \cite{halliquist2006ls}:
\begin{equation} \label{eq:gov_eq}
    \mathbf{M}\dot{\mathbf{v}} = \mathbf{f}_{\textrm{ext}} - \mathbf{f}_{\textrm{int}}
\end{equation}
where $\mathbf{M}$ is the mass matrix, $\dot{\mathbf{v}}$ is the acceleration, and $\mathbf{f}_{\textrm{ext}}$ and $\mathbf{f}_{\textrm{int}}$ are the external and internal forces respectively. The expressions for $\mathbf{M}$, $\mathbf{f}_{\textrm{ext}}$ and $\mathbf{f}_{\textrm{int}}$ are given as follow:
\begin{subequations}
\begin{align}
    &\mathbf{M} = \sum_{i=1}^m \int_{v_m}\rho \mathbf{N}_m^T \mathbf{N}_m dv \label{eq:M}\\
    &\mathbf{f}_{\textrm{ext}} = \sum_{i=1}^m \int_{v_m} \rho \mathbf{N}_m^T \mathbf{b} dv + \sum_{i=1}^m \int_{\partial b_1} \rho \mathbf{N}_m^T \mathbf{t} dS \label{eq:f_ext} \\
    &\mathbf{f}_{\textrm{int}} = \sum_{i=1}^m \int_{v_m} \mathbf{B}_m^T \boldsymbol{\sigma} dv \label{eq:f_int}
    \end{align}
\end{subequations}
where $\rho$ is the mass density and $\mathbf{N}_m$ are interpolation functions within element $m$ in \eqref{eq:M}. 
The external force $\mathbf{f}_{\textrm{ext}}$ in \eqref{eq:f_ext} comprises the body force vector $\mathbf{b}$, and the traction force boundary condition $\mathbf{t}$ on the boundary $\partial b_1$ . 
The internal force $\mathbf{f}_{\textrm{int}}$ in \eqref{eq:f_int} is related to the  stress tensor $\boldsymbol{\sigma}$ and the strain-displacement matrix $\mathbf{B}_m$. 
The surface and volume integrals appearing in \eqref{eq:M},~\eqref{eq:f_ext},~\eqref{eq:f_int} are numerically computed depending on the element type, and these integrals are summed over all $m$ elements.
The momentum balance equation \eqref{eq:gov_eq} are forward integrated in time, and the timestep is selected to satisfy the Courant-Friedrichs-Lewy (CFL) condition to ensure numerical stability.
\subsubsection{Contact Modeling}
Contact between deformable parts in nonlinear finite element analysis (FEA) is modeled as a constraint to the momentum balance equation given in \eqref{eq:gov_eq}, leading to the following modified balance equation:
\begin{equation} \label{eq:constrained_gov_eq}
     \mathbf{M}\dot{\mathbf{v}} + \mathbf{G}^{\textrm{T}}\bs{\lambda}_c= \mathbf{f}_{\textrm{ext}} - \mathbf{f}_{\textrm{int}}
\end{equation}
where $\mathbf{G}$ is the contact matrix and $\bs{\lambda}_c$ is the contact force, which appears as a lagrange multiplier in \eqref{eq:constrained_gov_eq}. The following geometric constraints are imposed:
\begin{subequations}
    \begin{align}
        & \mathbf{G}\mathbf{v}_i \le 0 \label{ineq:cond_impren}\\ 
        & \bs{\lambda}_{c,i} \ge 0  \label{ineq:pos_conforce} \\
        & \bs{\lambda}_{c,i}(\mathbf{G}\mathbf{v}_i) = 0  \label{eq:consistency} 
    \end{align}
\end{subequations}
The inequality in \eqref{ineq:cond_impren} indicates that the deformable parts do not penetrate each other, while the inequality in \eqref{ineq:pos_conforce} enforces that the contact force magnitude is always positive.
The equality constraint in \eqref{eq:consistency} represents the contact consistency condition. These constraints collectively form the Karush-Kuhn-Tucker (KKT) conditions. The solution to \eqref{eq:constrained_gov_eq} together with the constraint equations yields both the contact forces at the intersecting parts and the contact-enforced nodal kinematic solution \cite{zywicz1999general}. 

\subsubsection{Strain Calculations for Deformations}

Strain metrics have been reported to accurately describe soft-tissue injuries \cite{voycheck2010collagen,cloots2008biomechanics} and the associated risks of bio-implant failure \cite{albogha2016maximum}. This is particularly relevant for our application of grasping human arms, where understanding the strain distribution can help in preventing injury and optimizing the grasp. The FEM framework allows for direct computation of localized volumetric finite strain. The following displacement-strain relationships are utilized to compute the maximum/minimum principal strain ($e_{\textrm{max}}, e_{\textrm{min}}$) within the element:
\begin{subequations}
\begin{align} 
 & \mathbf{F}  = \mathbf{I} + \nabla_0\mathbf{u} {\label{eq:defgrad}} \\
 & \mathbf{E}  = \frac{1}{2}(\mathbf{F}^{\textrm{T}} \mathbf{F} - \mathbf{I}) = \sum_{i=1}^3 \frac{1}{2}({\lambda_{p,i}}^2 - 1)\mathbf{n}_i \otimes \mathbf{n}_i {\label{eq:green_lag}} \\
 &  e_i  = \lambda_{p,i} - 1 \quad \textrm{for } i=1,2,3 {\label{eq:principle_strains}} \\
 &  e_{\textrm{max}}  =\textrm{max}(e_i), \quad e_{\textrm{min}}=\textrm{min}(e_i) {\label{eq:mps}}
\end{align}
\end{subequations}
In \eqref{eq:defgrad}, the deformation gradient $\mathbf{F}$ is calculated from the displacement field $\mathbf{u}$. In \eqref{eq:green_lag}, the Green-Lagrange strain tensor $\mathbf{E}$ is computed, and its associated principal stretches $\lambda_{p,i}$ and principal directions $\mathbf{n}_i$ are obtained using an eigendecomposition. In \eqref{eq:principle_strains} and \eqref{eq:mps}, the principal strains and the maximum/minimum principal strains are calculated.

\subsection{The proposed integrative simulation framework}
We propose a novel integrative framework that benefits from both environments: (1) the computational convenience of rigid motion primitives in a multi-body dynamics simulator and (2) the high fidelity of soft contact mechanics in an FE solver. \par
A simplified system diagram is illustrated in Fig.~\ref{fig:rigid-vs-soft}~(e), where the robotic hand pose ($\mathbf{p}_h$, $R_h$) and joint trajectories ($\mathbf{q}_h$) from the rigid simulation environment are passed to the FE solver to compute high-fidelity biomechanical results. The \textit{Equilibrium State} is accomplished through rigid environment grasping simulation, described in \ref{subsec:grasp-in-mj}. 
Moving from the grasping \textit{Equilibrium State} to the \textit{Indented State}, FEM (as described in \ref{subsec:grasp-in-fem}) enables the robotic hand to penetrate soft materials, resulting in realistic deformation and a soft contact equilibrium grasp.\par

\begin{figure}[!h]
	\centering
	\includegraphics[width=1\columnwidth]{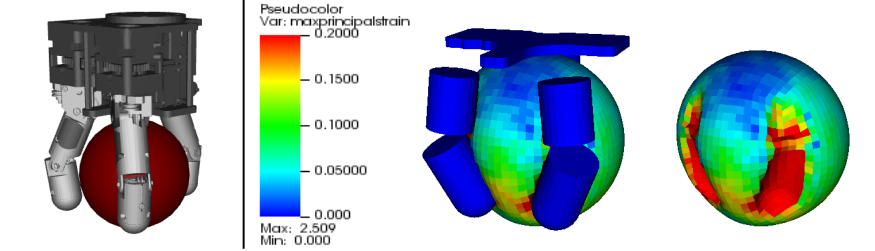}
	\caption{An illustrative example of a simulation done in the proposed integrative framework. A custom three-finger robotic hand grasps a rubber sphere. The left figure shows the \textit{Equilibrium State}, and the right two figures show the \textit{Indented State} with the maximum principal strain overlaid on the surface.}
	\label{fig:grasp-ball-qualitative-comparison}
\end{figure}

An illustrative example of using the proposed high-fidelity simulation framework is presented in Fig.~\ref{fig:grasp-ball-qualitative-comparison}: a robotic hand squeezing a rubber sphere. Figure~\ref{fig:grasp-ball-qualitative-comparison}~(left) is the \textit{Equilibrium State} with the rigid body assumption. Figure~\ref{fig:grasp-ball-qualitative-comparison}~(right) is the \textit{Indented State}, where surface deformation is demonstrated, and a heat map is overlaid to visualize the maximum principal strain. It can be seen that areas with greater deformation correspond to higher strain levels.\par

In the next section, building upon the simple toy problem of grasping and squeezing a rubber ball, we will further demonstrate two case studies of grasping and manipulating a human casualty within the proposed simulation framework.

\par

\section{Grasping Stability Analysis during Casualty Manipulation} \label{sec:grasping-stability-analysis}

Two case studies have been undertaken using the proposed integrative simulation framework. The objective is to thoroughly examine grasping stability and explore potential differences between simulations conducted in (i) a physics-based multi-body dynamics simulator and (ii) the proposed framework that integrates multi-body dynamics with FEM. Qualitative and quantitative comparisons are presented.\par

\begin{figure}[!h]
	\centering
	\vspace{5mm}
	\includegraphics[width=0.19\columnwidth]{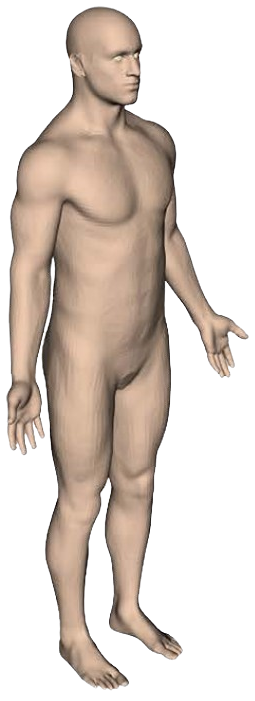}
	\includegraphics[width=0.196\columnwidth]{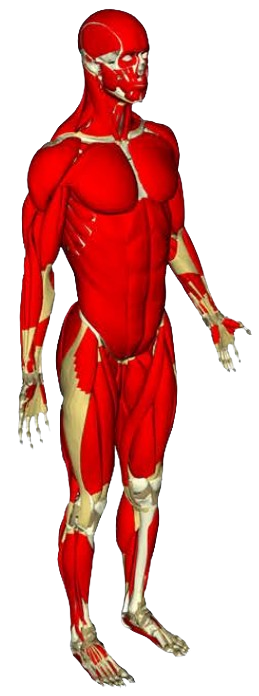}
	\includegraphics[width=0.185\columnwidth]{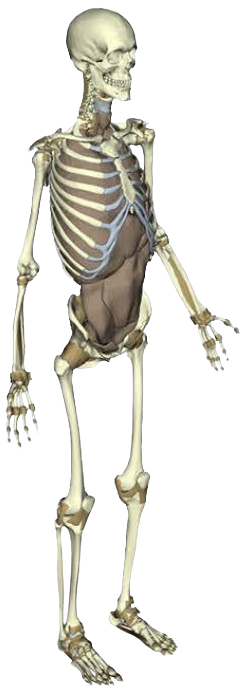}
	\includegraphics[width=0.18\columnwidth]{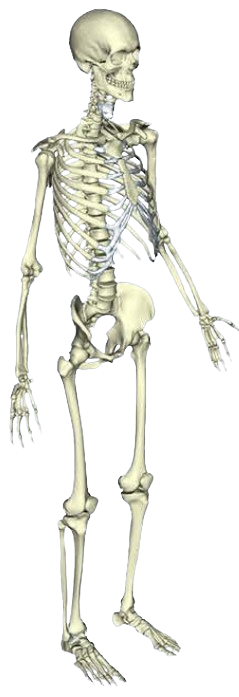}
	\caption{The Computational Anthropomorphic Virtual Experiment Man (CAVEMAN) model, developed by Corvid Technologies.}
	\label{fig:CAVEMAN}
\end{figure}

A gripper and a digital human model are used in all case studies in this work. The custom three-finger robotic hand shown in Fig.~\ref{fig:grasp-ball-qualitative-comparison}~(left) serves as the gripper. The Computational Anthropomorphic Virtual Experiment Man (CAVEMAN) model, depicted in Fig.~\ref{fig:CAVEMAN}, is employed as the high-fidelity digital human model. Developed by Corvid Technologies, it utilizes the detailed Zygote's 50th percentile male human CAD model \cite{Zygote}, which includes representations of all components of the human musculoskeletal system. Material models used in the CAVEMAN model have been derived from literature and experimental tissue testing. The model's connectivity matches all anatomical insertions as accurately as possible. The overall philosophy during the creation of the CAVEMAN human body model focused on minimizing simplifying assumptions to create an FE model that represents a cadaver as closely as possible. \par

In this study, the rigid body simulations are carried out using MuJoCo, while the FEM analysis is conducted with Velodyne \cite{Velodyne-Rajarshi}, a finite element solver developed by Corvid Technologies.\par

\par

\subsection{Case studies: grasp and manipulation of human body}
\subsubsection{Grasping at human upper limb} \label{sec: grasp-simulation}

\begin{figure}[!h]
	\centering
	\includegraphics[width=1\columnwidth]{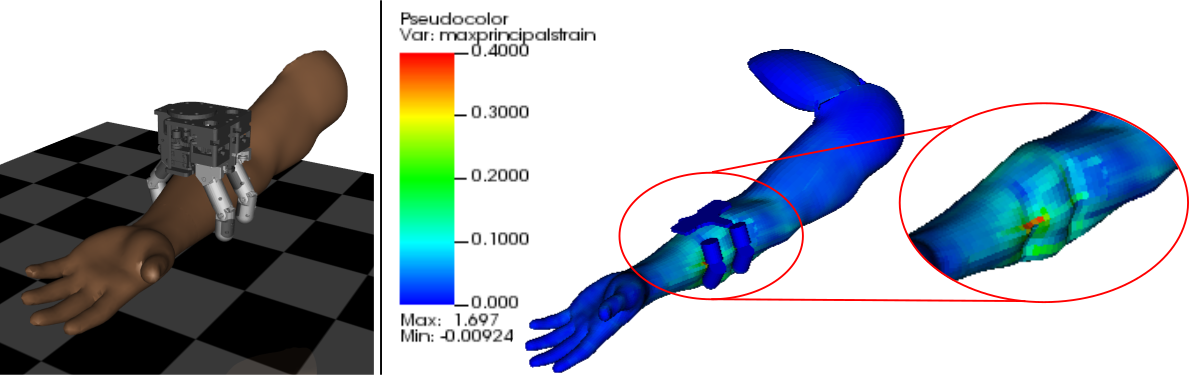}
	\caption{Case Study 1: Grasping the human upper limb using the proposed simulation framework. The left figure illustrates the simulation results in the rigid environment, and the right figure demonstrates grasping in the soft environment. The heat map depicts the maximum principal strain  $e_{\textrm{max}}$.}
	\label{fig:case-study-1}
\end{figure}

The first case study aims to realistically simulate the act of grasping the human body at the upper limb, as depicted in Fig.~\ref{fig:case-study-1}. We chose the upper limb as it aligns with our natural perception of grabbing someone. Our focus is on assessing grasping stability, in both rigid and soft simulation environments. \par

\par

\subsubsection{Human upper limb manipulation} \label{sec: grasp-and-pull-simulation}

\begin{figure}[!b]
	\centering
	\vspace{5mm}
	\includegraphics[width=0.99\columnwidth]{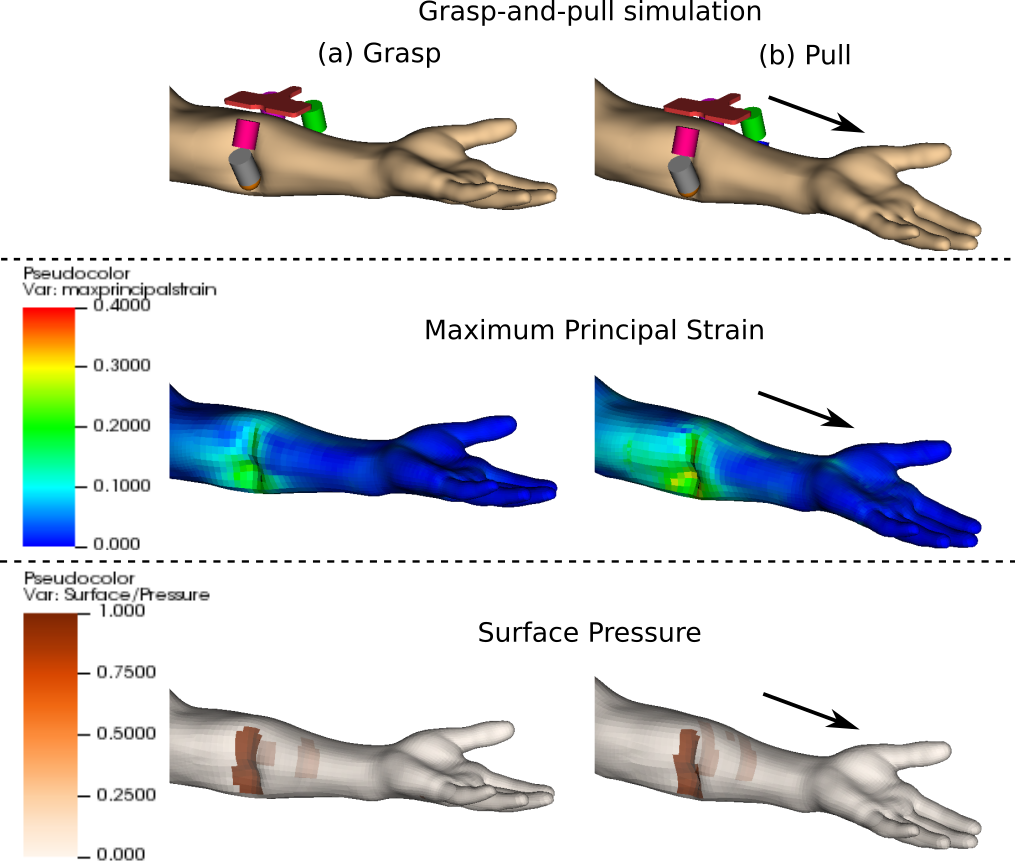}
	\caption{Case Study 2: Grasp-and-pull simulation using the proposed simulation framework. Firstly, the robotic hand closes its palm to grasp the CAVEMAN arm, as shown in Case Study 1. Then, the robotic hand pulls along the CAVEMAN arm towards the wrist by 2 cm. The three figures in column (a) depict the end of grasping, while the three figures in column (b) illustrate the end of pulling. Arrows indicate the pulling direction.}
	\label{fig:case-study-2}
\end{figure}

In the second case study, which follows the first, we command the robotic hand to pull along the arm direction after establishing a stable grasp. Here, we concentrate on exploring the stability of grasping in the context of manipulating the human body, mimicking a pulling motion. Figure~\ref{fig:case-study-2} illustrates the pulling of the CAVEMAN arm while maintaining the established grasp. \par

\par

\subsection{Qualitative Comparison}
Comparing simulation results between the rigid environment (the multi-body dynamics simulator) and the deformable environment (the proposed simulation framework) is visually straightforward. 
In the rigid setting, depicted in Fig.~\ref{fig:case-study-1}~(left), only several point contacts are established between the robotic hand and the human arm. 
Conversely, in the proposed simulation framework (deformable simulation environment) shown in Fig.~\ref{fig:case-study-1}~(right) and Fig.~\ref{fig:case-study-2},  realistic deformation can be observed in the human upper arm in the transition from the \textit{equilibrium state} to the \textit{indented state}. These deformations align with our expectations when grabbing someone's upper limb. Unlike in the rigid simulations, there is no penetration allowed even with aggressive force grasping on the human arm. The deformable environment permits more realistic results.\par

\begin{figure*}[!t]
	\centering
	\includegraphics[width=1.9\columnwidth]{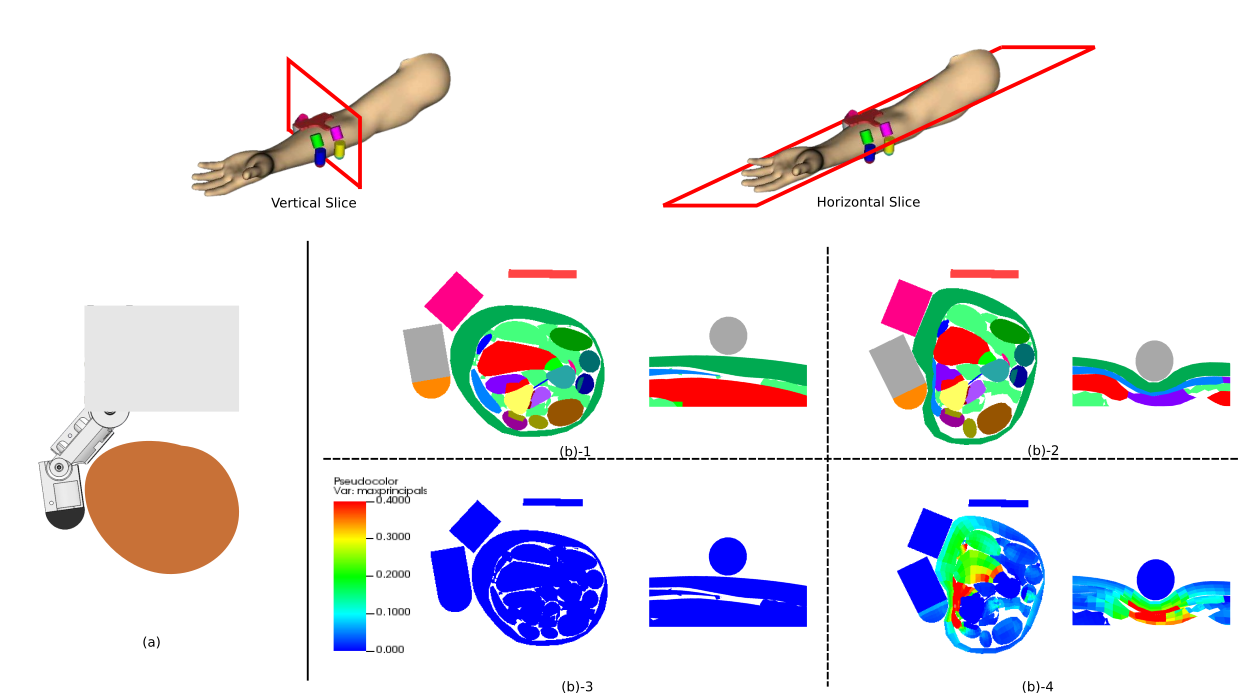}
	\caption{Grasping simulations in two engines: (a) in a multi-body dynamics simulator, and (b) in the proposed simulation framework. Sliced views are presented. (a) and (b)-left are sliced vertically, perpendicular to the human arm direction, as illustrated on the top left. This is referred to as the ``vertically-sliced view" in the paper content. (b)-right are sliced horizontally, parallel to the robotic hand palm, as illustrated on the top right. This is referred to as the ``horizontally-sliced view" in the paper content. (b)-1 and (b)-3 illustrate the \textit{equilibrium state}, and (b)-2 and (b)-4 depict the \textit{indented state}. Additionally, (b)-3 and (b)-4 include the maximum principal strain heat map overlay.}
	\label{fig:grasp_comparison_sliced_view}
\end{figure*}\par

\subsection{Quantitative Comparison of Grasping Stability}

Grasping stability is crucial in casualty manipulation, serving as a prerequisite for all human body manipulation tasks. Whether focusing on safety, precision, comfort, or adaptability, achieving a stable grasp is always necessary to proceed effectively. 
However, obtaining high-fidelity simulation results isn't always feasible, limiting the transition of casualty manipulation robot systems from conceptualization to real-world application.
In this section, we will evaluate grasping stability from two perspectives: (1) contact mechanisms and (2) prevention of grasping slippage. \par

Firstly, in comparing contact mechanisms in different simulation engines, the grasping simulation described in Case Study 1 (\ref{sec: grasp-simulation}) is executed in both a multi-body dynamics simulator and the proposed simulation framework. Results can be seen in Fig.~\ref{fig:grasp_comparison_sliced_view}. 
In Fig.~\ref{fig:grasp_comparison_sliced_view}~(a), the `vertically-sliced view' of the simulation result of grasping the digital human arm in the multi-body dynamics simulator is illustrated. Here, the rigid body assumption is employed, and therefore, only point contacts are allowed.
In Fig.~\ref{fig:grasp_comparison_sliced_view}~(b), the same grasping simulation executed in the proposed simulation framework is shown, with a mix of `vertically-sliced' and `horizontally-sliced' views to clearly illustrate contacts. Figure.~\ref{fig:grasp_comparison_sliced_view}~(b)-1 and (b)-3 are at the \textit{equilibrium state} and (b)-2 and (b)-4 are in the \textit{indented state}.\par

\begin{figure}[!b]
	\centering
	\includegraphics[width=0.99\columnwidth]{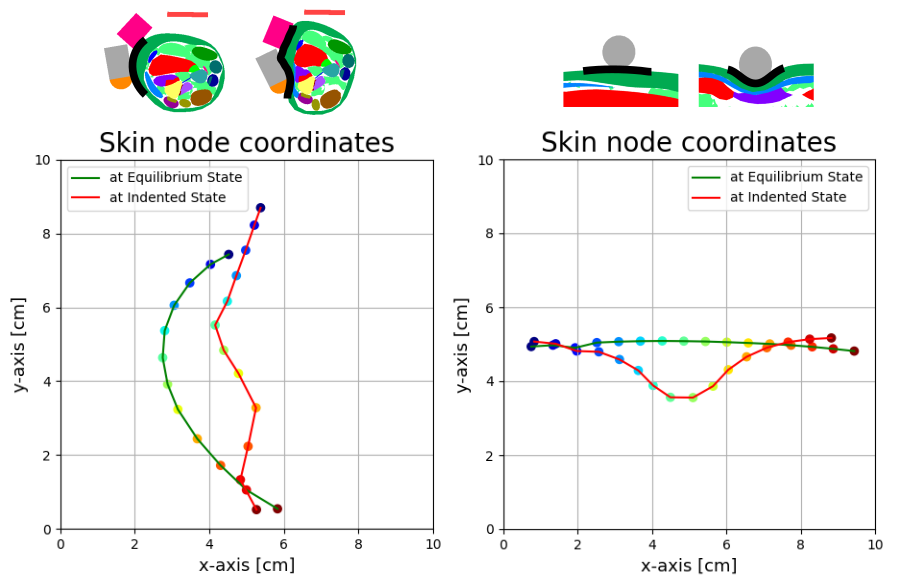}
	\caption{CAVEMAN arm node positions before and after grasping are shown by green and red lines, respectively, with vertically (left) and horizontally (right) sliced views for intuitive illustration.}
	\label{fig:skin_deformation_plots}
\end{figure}\par

The comparison in contact mechanisms is demonstrated, where the one with the rigid body assumption shows no deformation upon contact. Conversely, the result from the proposed simulation framework shows human body muscle deformation, enabled by the integrated FE engine. Skin surface deformation data have also been abstracted and plotted in Fig.~\ref{fig:skin_deformation_plots},
where the node positions before and after grasping are shown in the same plot for comparison and visualization.
Moreover, the interaction of body tissues resulting from the gripping motion is demonstrated in Fig.~\ref{fig:grasp_comparison_sliced_view}~(b)-3 and (b)-4 with the maximum principal strain heat map overlay. The maximum principal strain is generally proportional to the amount of deformation.
The results from the simulation in the FE engine realistically reflect the human body's biomechanical response compared to those in the rigid simulation environment. Additionally, these results are invaluable for further safety evaluations and injury assessments. \par

Secondly, one of the most important criteria in assessing grasping stability is the ability to prevent slippage when external disturbances are applied. 
For our specific application involving casualty manipulation tasks, we implemented grasp-and-pull simulations outlined in Case Study 2 (\ref{sec: grasp-and-pull-simulation}). These simulations are designed to assess grasping stability across various simulation engines, including the multi-body dynamics simulator used for the grasping simulations and the proposed simulation framework.\par

These two series of grasp-and-pull simulations involved defining 13 levels of grip tightness ranging from loose to tight. These levels were chosen to start from the \textit{equilibrium state} and gradually increase the grip force until reaching the gripping force level of the \textit{indented state}.
At each level of grip tightness, we instructed the robotic hand to move along the CAVEMAN arm towards the wrist by 2 cm while maintaining the grasp configuration. In the multi-body dynamics simulator, a one-dimensional slider is used to actuate the robotic hand, while in the proposed simulation framework, nodal displacement time history inputs are defined to achieve the same robotic hand pulling motion. \par 

Simulation results need to be compared at a similar grip tightness level to ensure the validity of the comparison. Grip tightness level was quantified by how tightly the fingers are closed around the CAVEMAN arm, meaning the normal contact force at each contact location. Given the grasping equilibrium condition, the normal contact force at each robotic hand finger should be at a similar level to preserve equilibrium. Therefore, we selected one of them—in our case, the thumb proximal link—to represent the grip tightness level. \par

\begin{figure}[!h]
	\centering
	\includegraphics[width=0.99\columnwidth]{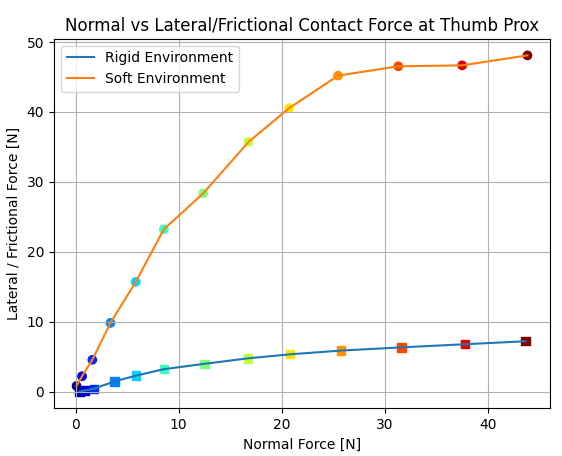}
	\caption{Comparison of normal and lateral contact forces during the grasp-and-pull simulations. The blue line represents results from the multi-body dynamics simulator, while the orange line represents results from the proposed simulation framework. Data points on the lines are color-coded to indicate similar normal contact forces.}
	\label{fig:normal_vs_lateral_contact_force}
\end{figure}\par

Figure~\ref{fig:normal_vs_lateral_contact_force} illustrates the results of two series of grasp-and-pull simulations. The focus is on comparing the contact normal force and lateral force. 
The blue line represents the contact information from the 13 multi-body dynamic simulations. The orange line represents the contact data from the proposed high-fidelity simulation framework. Data points are color-coded to correspond to the 13 levels of grip tightness. 
All contact forces are projected to their local contact frame,
\begin{align}
&\mathbf{f}_{n} = \left(\hat{\mathbf{n}}_s \;{\hat{\mathbf{n}}_s}^\top\right) \boldsymbol{\lambda}_c \\
&\mathbf{f}_{\mu} = \left(\mathbf{I} - \hat{\mathbf{n}}_s \; {\hat{\mathbf{n}}_s}^\top\right) \boldsymbol{\lambda}_c
\end{align}
where $\bs{\lambda}_c$ is the contact force vector, $\mathbf{f}_{n}$ and $\mathbf{f}_{\mu}$ are the normal and tangential projections of $\bs{\lambda}_c$, $\hat{\mathbf{n}}_s$ is a unit vector that represents the normal direction of the contact surface, and $\mathbf{I}$ is the identity matrix. \par

\begin{figure}[!h]
	\centering
	\includegraphics[width=0.85\columnwidth]{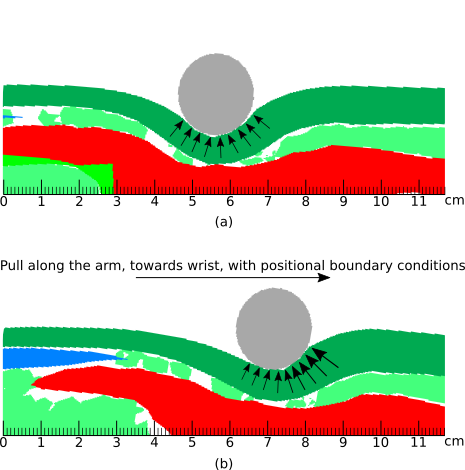}
	\caption{Human body deformation and reaction force during the grasp-and-pull simulation in the FE solver. (a) depicts the result at the end of grasping, while (b) illustrates the state at the end of pulling. }
	\label{fig:velodyne_grasp_arm_pull_contact}
\end{figure}

As observed from the blue line in Fig.~\ref{fig:normal_vs_lateral_contact_force}, the frictional/normal force ratio is low and close to a linear relation. This is attributed to the rigid body point contact assumption. This assumption results in a contact mechanism where the contact force is more accurate in the normal direction. However, it does not effectively simulate the realism in preventing lateral motion, leading to grasp slippage.
Conversely, the contact profile from the proposed high-fidelity simulation framework behaves differently. A data point on the steeper line (orange line) in Fig.~\ref{fig:normal_vs_lateral_contact_force} indicates a grasp that offers better resistance to lateral motion, indicating a more stable grasp.
On the orange line, the lateral force is consistently greater than the normal contact force, regardless of the grip tightness level. This is because, in the proposed simulation framework, not only does the frictional force contribute to the lateral force, but more importantly, the contact force at deformation prevents the CAVEMAN arm from slipping when disturbances are applied. Figure~\ref{fig:velodyne_grasp_arm_pull_contact} illustrates the elastic deformation that prevents grasping slippage. \par

\par

The combination of FEM and the detailed construction of the CAVEMAN model in the proposed simulation framework enables the attainment of high-fidelity contact results. FEM effectively breaks down complex problems into smaller, more manageable elements interconnected by nodes. Each element is characterized by its material and geometric properties, facilitating accurate calculations of its behavior under diverse conditions. Notably, FEM excels in simulating and analyzing continuous media with a focus on internal interactions.
Similarly, the intricate responses of under-skin body tissues to human-robot physical interactions can benefit from the unique features of FEM. Human body tissues exhibit complex properties that vary with factors like location, orientation, and strain rate \cite{body_tissue_definition}. The CAVEMAN model stands as a comprehensive representation of the human body, encompassing the full skeleton, 397 muscles, 342 ligaments, 16 organs, and skin. This extensive detail ensures an accurate representation of various biological tissues. \par

We apply the strength of FEM to simulating robotic grasping and manipulation tasks. We believe that only using multi-body dynamics simulators is insufficient for supporting casualty manipulation tasks. As discussed previously, the rigid point contact assumption makes the simulation results deviate significantly from reality. Only limited material properties, such as friction, can be specified in a multi-body dynamics simulator. 
However, in the proposed simulation framework with FEM integration, material properties are highly customizable and can be defined very close to human biological systems. As a result, this capability enables the representation of complex nonlinear biomechanical behaviors that occur during physical contact between humans and robots.\par

In summary, our study highlights a significant contribution by quantitatively revealing the biases and inaccuracies inherent in current SotA physics-based multi-body dynamics simulators when predicting grasping stability in casualty manipulation scenarios. 
For instance, as illustrated in Fig.~\ref{fig:normal_vs_lateral_contact_force},  the multi-body dynamics simulator predicts a frictional drag force of less than 10 N, even with a high grasping actuation force of 40 N. Meanwhile, with the same level of grip tightness, the proposed high-fidelity simulation framework suggests a lateral drag force of almost 50 N.
The trend line of the proposed simulation framework has a linear region and reaches a plateau at higher grip tightness levels; however, the non-realistic results from the multi-body dynamics simulator show a completely linear trend line. This substantial discrepancy arises due to the differing fidelity levels of the two tools, each with varying computational costs. It is crucial for robotic planners to be aware of these potential unrealistic predictions by SotA simulators. Future research may explore a more balanced solution between accuracy and computational cost. \par

\subsection{Comparisons on Computation Time and Environments}
When setting up simulations, the approaches and considerations differ between a multi-body dynamics simulator and the proposed framework. Multi-body dynamics simulators focus on defining the rigid bodies and their interconnected joints or constraints, emphasizing motion and interaction over material properties. While material properties are not directly modeled, parameters such as friction coefficient and contact stiffness are defined for accurate interactions between the robotic hand and human model.
Conversely, in the proposed high-fidelity simulation framework with FEM integrated, CAVEMAN and the robotic hand are discretized into small elements connected by nodes, allowing for a detailed representation of geometry and material properties. Specific materials must be defined for each human body part and the robotic hand. In addition, boundary conditions and material properties need to be specified for each element.\par

When comparing simulation time consumption and computational cost, multi-body dynamics simulators typically offer nearly real-time simulations and lower computational costs compared to the proposed simulation framework. This is because the former focuses on rigid body motion and interaction. In contrast, the latter requires longer time and higher computational cost due to fine meshing and the complex nature of FE models, especially for large-scale systems such as a high-fidelity digital human model with nonlinear behavior. Each simulation of grasping at the human upper limb, as shown in Fig.~\ref{fig:case-study-1}, needs over 10 hours to compute. \par

\section{Discussion}

The proposed high-fidelity simulation framework for casualty manipulation represents a novel and effective integration of physics-based multi-body dynamics simulators and FEM. We are not utilizing the mechanics and physics results from the physics-based multi-body dynamics simulator; rather, we are only leveraging its effectiveness in solving robot kinematics. Our study suggests that relying solely on the physics-based multi-body dynamics simulator may not be adequate for realistically simulating casualty manipulation, as demonstrated in the findings presented in \ref{sec: grasp-and-pull-simulation}.\par



\section{Conclusion}

In conclusion, this paper has addressed critical limitations in the development of robot-assisted casualty manipulation by introducing a novel simulation framework and conducting comprehensive investigations on grasping stability. The integration of FEM into the simulation pipeline has enabled accurate modeling of biomechanical reactions during robot manipulation actions, leveraging a high-fidelity digital human model. Through qualitative and quantitative comparisons across various simulation engines, the necessity and superior performance of the proposed framework have been validated. This work marks an important initial step toward realizing a feasible solution for robot casualty manipulation, laying a solid foundation for future advancement in this field. \par

To advance this work towards real-world applicability, several future research directions can be pursued. 
Firstly, while FEM shows promise, its computational demands are a concern, especially for field deployment where real-time or near real-time operation is crucial. Our future plans include leveraging various machine learning techniques to train models capable of predicting human body biomechanical reactions during robot contact. This approach aims to accelerate the response time of the casualty manipulation robot system, bringing it closer to real-world deployment.
Secondly, the risk of causing further injuries to wounded soldiers during grasping and manipulation actions is another significant obstacle that limits the rapid development and iteration of casualty manipulation robot systems. Leveraging our high-fidelity digital human model, we intend to investigate the correlation between robot actions and potential human body injuries, such as exceeding joint load limits and ligament tears. This exploration will guide the development of biomechanically-informed, safer, and more effective casualty extraction robot systems.\par

\section*{Acknowledgments}

This material is based on work that was supported by the U.S. Army Medical Research and Material Command (USAMRDC), Telemedicine and Advanced Technology Research Center (TATRC) through the Small Business Innovation research (SBIR) Program under contract number W81XWH22P0030. The views, opinions and/or findings contained in this research/presentation/publication are those of the author(s)/company and do not necessarily reflect the views of the Department of Defense and should not be construed as an official DoD/Army position, policy or decision unless so designated by other documentation.  No official endorsement should be made. Reference herein to any specific commercial products, process, or service by trade name, trademark, manufacturer, or otherwise, does not necessarily constitute or imply its endorsement, recommendation, or favoring by the U.S. Government.

\bibliographystyle{IEEEtran}
\bibliography{bib/IEEEabrv,bib/related_work}

\begin{thebibliography}{10}
\providecommand{\url}[1]{#1}
\csname url@samestyle\endcsname
\providecommand{\newblock}{\relax}
\providecommand{\bibinfo}[2]{#2}
\providecommand{\BIBentrySTDinterwordspacing}{\spaceskip=0pt\relax}
\providecommand{\BIBentryALTinterwordstretchfactor}{4}
\providecommand{\BIBentryALTinterwordspacing}{\spaceskip=\fontdimen2\font plus
\BIBentryALTinterwordstretchfactor\fontdimen3\font minus
  \fontdimen4\font\relax}
\providecommand{\BIBforeignlanguage}[2]{{%
\expandafter\ifx\csname l@#1\endcsname\relax
\typeout{** WARNING: IEEEtran.bst: No hyphenation pattern has been}%
\typeout{** loaded for the language `#1'. Using the pattern for}%
\typeout{** the default language instead.}%
\else
\language=\csname l@#1\endcsname
\fi
#2}}
\providecommand{\BIBdecl}{\relax}
\BIBdecl

\bibitem{adams2013robotic}
J.~A. Adams, ``Robotic technologies for first response: A review,'' in
  \emph{Handbook of Emergency Response}.\hskip 1em plus 0.5em minus 0.4em\relax
  CRC Press, 2013, pp. 38--69.

\bibitem{adams2024human}
J.~A. Adams, J.~Scholtz, and A.~Sciarretta, ``Human--robot teaming challenges
  for the military and first response,'' \emph{Annual Review of Control,
  Robotics, and Autonomous Systems}, vol.~7, 2024.

\bibitem{murphy2007search}
R.~R. Murphy, ``Search and rescue robotics,'' \emph{Springer handbook of
  robotics}, pp. 1151--1173, 2007.

\bibitem{Bhatia2011}
S.~Bhatia, H.~S. Dhillon, and N.~Kumar, ``Alive human body detection system
  using an autonomous mobile rescue robot,'' in \emph{2011 Annual IEEE India
  Conference}, 2011, pp. 1--5.

\bibitem{Niroui2019}
F.~Niroui, K.~Zhang, Z.~Kashino, and G.~Nejat, ``Deep reinforcement learning
  robot for search and rescue applications: Exploration in unknown cluttered
  environments,'' \emph{IEEE Robotics and Automation Letters}, vol.~4, no.~2,
  pp. 610--617, 2019.

\bibitem{Lygouras2019}
\BIBentryALTinterwordspacing
E.~Lygouras, N.~Santavas, A.~Taitzoglou, K.~Tarchanidis, A.~Mitropoulos, and
  A.~Gasteratos, ``Unsupervised human detection with an embedded vision system
  on a fully autonomous uav for search and rescue operations,'' \emph{Sensors},
  vol.~19, no.~16, 2019. [Online]. Available:
  \url{https://www.mdpi.com/1424-8220/19/16/3542}
\BIBentrySTDinterwordspacing

\bibitem{Cardona2019}
\BIBentryALTinterwordspacing
G.~A. Cardona and J.~M. Calderon, ``Robot swarm navigation and victim detection
  using rendezvous consensus in search and rescue operations,'' \emph{Applied
  Sciences}, vol.~9, no.~8, 2019. [Online]. Available:
  \url{https://www.mdpi.com/2076-3417/9/8/1702}
\BIBentrySTDinterwordspacing

\bibitem{Deng2020}
W.~Deng, K.~Huang, X.~Chen, Z.~Zhou, C.~Shi, R.~Guo, and H.~Zhang, ``Semantic
  rgb-d slam for rescue robot navigation,'' \emph{IEEE Access}, vol.~8, pp.
  221\,320--221\,329, 2020.

\bibitem{Queralta2020}
J.~P. Queralta, J.~Taipalmaa, B.~Can~Pullinen, V.~K. Sarker, T.~Nguyen~Gia,
  H.~Tenhunen, M.~Gabbouj, J.~Raitoharju, and T.~Westerlund, ``Collaborative
  multi-robot search and rescue: Planning, coordination, perception, and active
  vision,'' \emph{IEEE Access}, vol.~8, pp. 191\,617--191\,643, 2020.

\bibitem{Cruz2021}
C.~Cruz, G.~Sánchez, A.~Barrientos, and J.~Cerro, ``Autonomous thermal vision
  robotic system for victims recognition in search and rescue missions,''
  \emph{Sensors}, vol.~21, p. 7346, 11 2021.

\bibitem{micheal_yip_contactless_weight}
J.~Lee, E.~Quist, J.~Chambers, M.~Yip, and N.~Fisher, ``Contactless weight
  estimation of human body and body parts for safe robotics-assisted casualty
  extraction,'' in \emph{2023 IEEE/RSJ International Conference on Intelligent
  Robots and Systems (IROS)}, 2023, pp. 6550--6556.

\bibitem{ValkyrieSBIR}
``Valkyrie: A patient recovery robot,''
  \url{https://www.sbir.gov/sbirsearch/detail/203536}, accessed: 2023-02-06.

\bibitem{REXREVsbir}
M.~K.~B. Gary R.~Gilbert, ``Us department of defense research in robotic
  unmanned systems for combat casualty care,'' 2010.

\bibitem{BookSurgicalRobotics}
R.~M.~S. Jacob~Rosen, Black~Hannaford, \emph{Surgical Robotics: Systems
  Applications and Visions}.\hskip 1em plus 0.5em minus 0.4em\relax New York,
  USA: Springer, 2011, ch.~2, pp. 13--32.

\bibitem{BEAR}
``Bear,'' \url{https://robotsguide.com/robots/bear}, accessed: 2024-02-23.

\bibitem{Sun2022}
Z.~Sun, H.~Yang, Y.~Ma, X.~Wang, Y.~Mo, H.~Li, and Z.~Jiang, ``Bit-dmr: A
  humanoid dual-arm mobile robot for complex rescue operations,'' \emph{IEEE
  Robotics and Automation Letters}, vol.~7, no.~2, pp. 802--809, 2022.

\bibitem{Yim2009}
M.~Yim, T.~Cragg, and S.-K. Hayat, ``Towards small robot aided victim
  manipulation,'' in \emph{2009 IEEE International Workshop on Safety, Security
  and Rescue Robotics (SSRR 2009)}, 2009, pp. 1--6.

\bibitem{yim2011towards}
M.~Yim and J.~Laucharoen, ``Towards small robot aided victim manipulation,''
  \emph{Journal of Intelligent \& Robotic Systems}, vol.~64, pp. 119--139,
  2011.

\bibitem{williams2017analysis}
A.~Williams, W.~Saab, and P.~Ben-Tzvi, ``Analysis of differential mechanisms
  for a robotic head stabilization system,'' in \emph{International Design
  Engineering Technical Conferences and Computers and Information in
  Engineering Conference}, vol. 58189.\hskip 1em plus 0.5em minus 0.4em\relax
  American Society of Mechanical Engineers, 2017, p. V05BT08A008.

\bibitem{williams2019robotic}
A.~Williams, B.~Sebastian, and P.~Ben-Tzvi, ``A robotic head stabilization
  device for medical transport,'' \emph{Robotics}, vol.~8, no.~1, p.~23, 2019.

\bibitem{micheal_yip}
E.~Peiros, Z.-Y. Chiu, Y.~Zhi, N.~Shinde, and M.~C. Yip, ``Finding
  biomechanically safe trajectories for robot manipulation of the human body in
  a search and rescue scenario,'' in \emph{2023 IEEE/RSJ International
  Conference on Intelligent Robots and Systems (IROS)}, 2023, pp. 167--173.

\bibitem{FEM_soft_tissue_liver}
A.~Idkaidek and J.~Iwona, ``Towarad high-speed 3d nonlinear soft tissue
  deformation simulations using abaqus software,'' in \emph{Journal of Robotic
  Surgery}, vol.~9, 2015, pp. 299--310.

\bibitem{Palmeri2010}
M.~Palmeri, A.~Sharma, R.~Bouchard, R.~Nightingale, and K.~Nightingale, ``A
  finite-element method model of soft tissue response to impulsive acoustic
  radiation force,'' \emph{IEEE Transactions on Ultrasonics, Ferroelectrics,
  and Frequency Control}, vol.~52, no.~10, pp. 1699--1712, 2005.

\bibitem{FEM_soft_tissue_survey_2005}
U.~Meier, O.~Lopez, C.~Monserrat, M.-C. Juan, and M.~Alcañiz~Raya, ``Real-time
  deformable models for surgery simulation: A survey,'' \emph{Computer methods
  and programs in biomedicine}, vol.~77, pp. 183--97, 04 2005.

\bibitem{FEM_soft_tissue_survey_2019}
\BIBentryALTinterwordspacing
J.~Zhang, Y.~Zhong, and C.~Gu, ``Deformable models for surgical simulation: A
  survey,'' \emph{IEEE Reviews in Biomedical Engineering}, vol.~11, pp.
  143--164, 2019. [Online]. Available:
  \url{https://api.semanticscholar.org/CorpusID:50785173}
\BIBentrySTDinterwordspacing

\bibitem{MatieGraspWithSoftFingerDeformation}
M.~Ciocarlie, A.~Miller, and P.~Allen, ``Grasp analysis using deformable
  fingers,'' in \emph{2005 IEEE/RSJ International Conference on Intelligent
  Robots and Systems}, 2005, pp. 4122--4128.

\bibitem{IPC_GraspSim}
C.~M. Kim, M.~Danielczuk, I.~Huang, and K.~Goldberg, ``Ipc-graspsim: Reducing
  the sim2real gap for parallel-jaw grasping with the incremental potential
  contact model,'' in \emph{2022 International Conference on Robotics and
  Automation (ICRA)}, 2022, pp. 6180--6187.

\bibitem{Xu2021TRO}
J.~Xu, T.~Aykut, D.~Ma, and E.~Steinbach, ``6dls: Modeling nonplanar frictional
  surface contacts for grasping using 6-d limit surfaces,'' \emph{IEEE
  Transactions on Robotics}, vol.~37, no.~6, pp. 2099--2116, 2021.

\bibitem{howell2013compliant}
L.~L. Howell, \emph{Compliant mechanisms}.\hskip 1em plus 0.5em minus
  0.4em\relax Springer, 2013.

\bibitem{faybook}
R.-F. Fay, \emph{Flexible Bars}.\hskip 1em plus 0.5em minus 0.4em\relax
  Butterworths, 1962.

\bibitem{halliquist2006ls}
J.~Halliquist, ``Ls-dyna theory manual,'' \emph{California: Livermore Software
  Technology Corporation}, 2006.

\bibitem{zywicz1999general}
E.~Zywicz and M.~A. Puso, ``A general conjugate-gradient-based
  predictor--corrector solver for explicit finite-element contact,''
  \emph{International journal for numerical methods in engineering}, vol.~44,
  no.~4, pp. 439--459, 1999.

\bibitem{voycheck2010collagen}
C.~A. Voycheck, P.~J. McMahon, and R.~E. Debski, ``Collagen fiber alignment and
  maximum principle strain in the axillary pouch predict location of failure
  during uniaxial extension,'' in \emph{Summer Bioengineering Conference}, vol.
  44038.\hskip 1em plus 0.5em minus 0.4em\relax American Society of Mechanical
  Engineers, 2010, pp. 463--464.

\bibitem{cloots2008biomechanics}
R.~Cloots, H.~Gervaise, J.~Van~Dommelen, and M.~Geers, ``Biomechanics of
  traumatic brain injury: influences of the morphologic heterogeneities of the
  cerebral cortex,'' \emph{Annals of biomedical engineering}, vol.~36, pp.
  1203--1215, 2008.

\bibitem{albogha2016maximum}
M.~H. Albogha, T.~Kitahara, M.~Todo, H.~Hyakutake, and I.~Takahashi, ``Maximum
  principal strain as a criterion for prediction of orthodontic mini-implants
  failure in subject-specific finite element models,'' \emph{The Angle
  Orthodontist}, vol.~86, no.~1, pp. 24--31, 2016.

\bibitem{Zygote}
\BIBentryALTinterwordspacing
Zygote. Solid 3d male model. [Online]. Available:
  \url{https://www.zygote.com/cad-models/collections-products/solid-3d-male-collection}
\BIBentrySTDinterwordspacing

\bibitem{Velodyne-Rajarshi}
R.~Roy, C.~M. Spurlock, K.~D. Butz, and K.~Lister, ``Incorporating hierarchical
  soft-tissue failure in whole body finite element models,'' in \emph{2020
  International Research Council on the Biomechanics of Injury}, 2020, pp.
  217--218.

\bibitem{body_tissue_definition}
\BIBentryALTinterwordspacing
``Body tissues | seer training.'' [Online]. Available:
  \url{https://training.seer.cancer.gov/anatomy/cells_tissues_membranes/tissues/}
\BIBentrySTDinterwordspacing

\end{thebibliography}

\balance
\end{document}